\documentclass[10pt,twocolumn,letterpaper]{article}

\usepackage{cvpr}              

\usepackage{graphicx}
\usepackage{amsmath}
\usepackage{amssymb}
\usepackage{booktabs}

\usepackage{amsmath}
\usepackage{amssymb}
\usepackage{mathtools}
\usepackage{amsthm}
\usepackage{bbm}

\usepackage{multirow}
\usepackage{adjustbox}
\usepackage{textcomp}
\usepackage{gensymb}
\usepackage{wrapfig}
\usepackage{setspace}
\usepackage[dvipsnames]{xcolor}
\usepackage[ruled,linesnumbered, noend]{algorithm2e}
\def \HiLi{\leavevmode\rlap{\hbox to \hsize{\color{yellow!50}\leaders\hrule height .8\baselineskip depth .5ex\hfill}}}
\usepackage{hyperref}
\usepackage[capitalize,noabbrev]{cleveref}
\theoremstyle{plain}

\theoremstyle{definition}

\theoremstyle{remark}

\usepackage[ruled,linesnumbered,noend]{algorithm2e}
\usepackage{xspace}

\newcommand{\sys}{DRAM\xspace}

\newcommand{\jc}[2]{{\textcolor{blue}{\bf [JC: #2]}}}

\usepackage[capitalize]{cleveref}
\crefname{section}{Sec.}{Secs.}
\Crefname{section}{Section}{Sections}
\Crefname{table}{Table}{Tables}
\crefname{table}{Tab.}{Tabs.}

\def\cvprPaperID{22} 
\def\confName{CVPR}
\def\confYear{2023}

\begin{document}

\title{Test-time Detection and Repair of Adversarial Samples via Masked Autoencoder}

\author{Yun-Yun Tsai$^1$~~~Ju-Chin Chao$^1$~~~Albert Wen$^1$~~~Zhaoyuan Yang$^2$ \\~~~Chengzhi Mao$^1$~~~Tapan Shah$^2$~~~Junfeng Yang$^1$ \\
$^1$Columbia University~~~$^2$GE Research\\
{\tt\small \{yt2781,jc5859,aw3575,cm3797\}@columbia.edu,}\\
{\tt\small\{zhaoyuan.yang,tapan.shah\}@ge.com,}
{\tt\small junfeng@cs.columbia.edu}
}



\maketitle

\begin{abstract}
Training-time defenses, known as adversarial training, incur high training costs and do not generalize to unseen attacks. Test-time defenses solve these issues but most existing test-time defenses require adapting the model weights, therefore they do not work on frozen models and complicate model memory management. The only test-time defense that does not adapt model weights aims to adapt the input with self-supervision tasks. However, we empirically found these self-supervision tasks are not sensitive enough to detect adversarial attacks accurately. In this paper, we propose \sys, a novel defense method to detect and repair adversarial samples at test time via Masked autoencoder (MAE).
We demonstrate how to use MAE losses to build a Kolmogorov-Smirnov test to detect adversarial samples. Moreover, we use the MAE losses to calculate input reversal vectors that repair adversarial samples resulting from previously unseen attacks.
Results on large-scale ImageNet dataset show that, compared to all detection baselines evaluated, \sys achieves the best detection rate (82\% on average) on all eight adversarial attacks evaluated. For attack repair, \sys improves the robust accuracy by 6\%$\sim$41\% for standard ResNet50 and 3\%$\sim$8\% for robust ResNet50 compared with the baselines that use contrastive learning and rotation prediction.


\end{abstract}


\section{Introduction}\label{sec:introduction}




\begin{figure}
    \centering
    \includegraphics[scale=0.70]{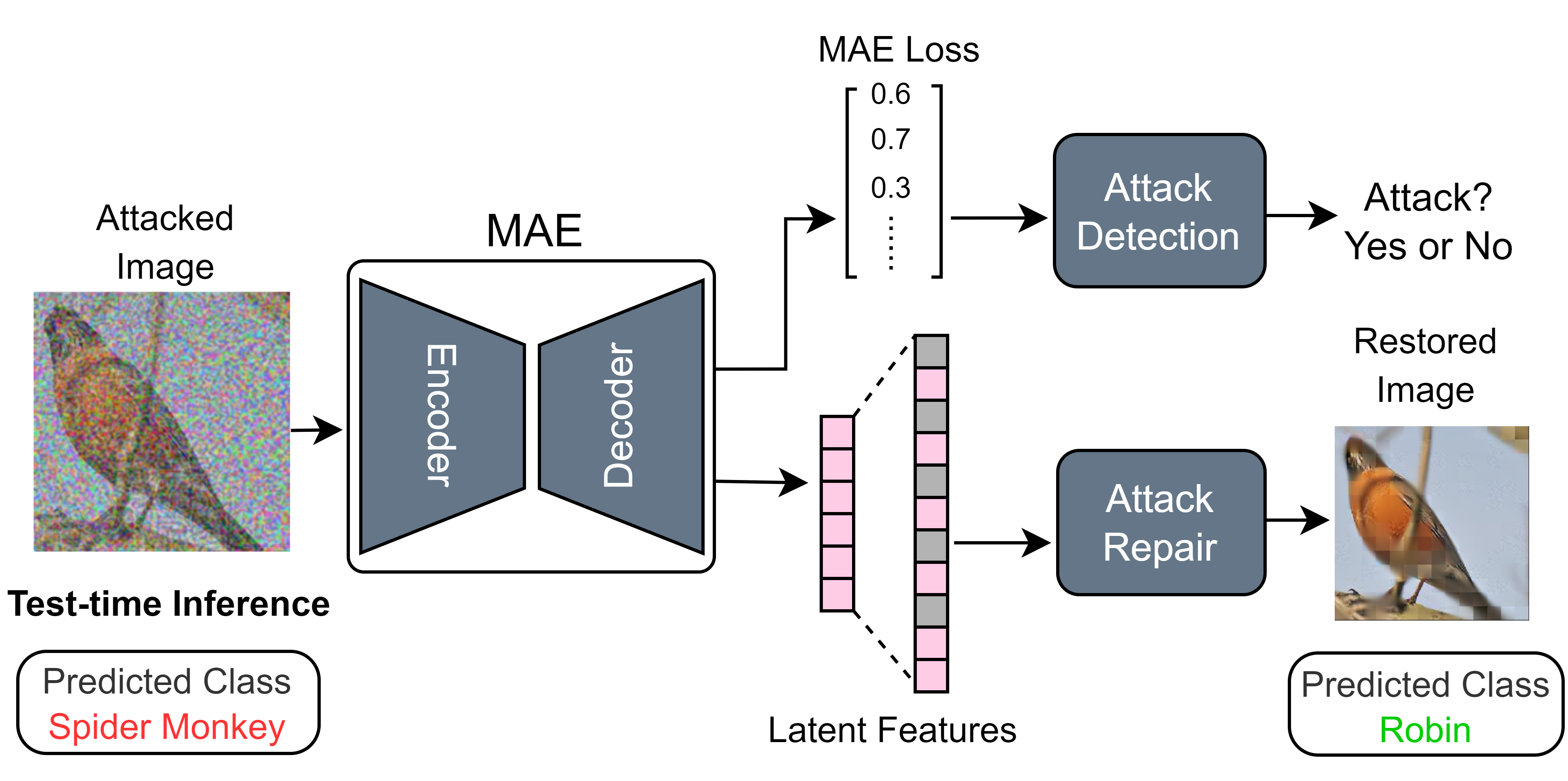}
    \vspace{-5mm}
    \caption{A schematic of \sys. (1) Attack detection: Leverages the MAE loss to detect adversarial samples via a statistic-based test. (2) Attack Repair: Adapts inputs by adding a reversal vector to the input space and optimizing it with MAE loss, which can potentially repair the adversarial sample.}
    \label{fig:flow}
\end{figure}

Despite remarkable advances in prediction accuracy and wide adoption of deep learning (DL) models, they are known to be vulnerable to adversarial attacks in which attackers add carefully crafted, human-imperceptible perturbations to the samples to fool the models~\cite{goodfellow2014explaining,carlini2017towards,chen2017zoo,madry2017towards}.
Most prior defenses happen at training time~\cite{kurakin2016adversarial, madry2018towards, zhang2019theoretically}, which incurs high training cost and cannot defend against unseen attacks. Recent work proposed test-time defenses~\cite{zhou2020deep, wang2021tent, croce2022evaluating, gandelsman2022test} that adapt the model weights to improve prediction accuracy on out-of-distribution data, but they require access to the weights and cannot work on frozen models. Additionally, they may require multiple versions of the same models to operate simultaneously, complicating model management and increasing memory consumption. Another test-time defense~\cite{mao2021adversarial, tsai2023self} reverses attacks by adapting inputs computed using self-supervised learning (SSL) tasks such as contrastive learning~\cite{simclr} or rotation prediction~\cite{gidaris2018unsupervised}. However, as our experiments reveal, these tasks are too weak to capture sufficiently detailed representations for effective detection of adversarial samples.

This paper presents \sys, a test-time defense using masked autoencoder (MAE)~\cite{MaskedAutoencoders2021},  one of the strongest SSL tasks, to detect and repair adversarial attacks, including unforeseen ones. MAE is a ViT-based architecture that reconstructs masked patches based on the sparse information of unmasked patches. Prior work~\cite{MaskedAutoencoders2021} showed that MAE is both efficient at reducing redundancy in feature representations and capturing detailed information from holistic image statistics.
We leverage MAE to detect adversarial samples generated from potentially unforeseen attacks via a statistic-based test on the MAE loss. Moreover, without knowing the ground truth of the samples, \sys repairs the samples by minimizing the MAE loss at test time. We highlight our main contribution as follows:

\begin{itemize}
    \item We propose a test-time defense method for unknown adversarial attacks using self-supervised objectives and masked autoencoder.

    \item We illustrate the method of \sys on both detection and repair: (1) Adversarial detection by statistic-based learning with reconstruction loss. (2) Repair of the adversarial samples by injecting imperceptible perturbations and optimizing with self-supervised loss. (3) We launch an adaptive attack that assumes the attacker has knowledge of our defense method and shows the effectiveness of \sys on the defense-aware attack.
    
    \item \sys achieves 82\% detection rate on eight kinds of adversarial samples average and outperforms other detection baselines. For attack repair, \sys successfully repairs the adversarial samples and improves robust accuracy by 6\%$\sim$41\% on eight attacks for ResNet50 under normal training setting and 3\%$\sim$8\% under robust training setting. Our method outperforms two self-supervision tasks, including contrastive learning and rotation prediction.
    

\end{itemize}


\section{Related Works}\label{sec:related}
\paragraph{Self-Supervised Learning}
Self-supervised learning (SSL) is one of the most efficient ways to learn effective representations from images without annotations~\cite{de1994learning,chen2020improved, NEURIPS2020_70feb62b, hendrycks2019using}. Prior works have shown that representations learned from different pretext tasks (e.g., jigsaw puzzles~\cite{noroozi2016unsupervised}, rotation prediction~\cite{gidaris2018unsupervised}, image colorization~\cite{larsson2016learning} and deep clustering~\cite{ji2019invariant}) can be leveraged on several downstream tasks such as image classification~\cite{chen2020simple}, object detection~\cite{doersch2015unsupervised} and test-time domain adaptation~\cite{pmlr-v119-sun20b}. Another well-known branch of SSL is contrastive learning, which aims at grouping associated features for a set of transformations for samples and distancing from other samples with dissimilar features in the dataset~\cite{chen2020big, he2020momentum, park2020contrastive}. Prior works proposed to use SSL method for outlier detection~\cite{hendrycks2019using, sehwag2021ssd, zeng2021adversarial}, which aims at learning generalizable out-of-detection features and rejecting them during the testing time. However, these discriminative SSL tasks learn to discard a large portion of their input information. Representations learned from these tasks keep only a small amount of information or features from the input space. Consequently, these objective functions are less sensitive to adversarial attacks.





\paragraph{Masked Autoencoder}
MAE learns to reconstruct the randomly masked patches of a given input. Recent research leverages MAE to learn downstream tasks since it allows the models to learn representations beyond the low-level statistics. It first encodes a small subset of unmasked patches to latent space as embedding. Then, the masked tokens are inserted after encoding. The concatenation of unmasked patches' embedding and masked tokens are processed by the decoder to predict the unmasked patches~\cite{MaskedAutoencoders2021, xie2022simmim}. Existing works show the success of mask-based pre-training on different modalities, such as improving point cloud models for 3D object detection~\cite{Voxel-MAE}, and model adaptation on sequential data~\cite{wang2021tsdae}. In ~\cite{gong2022reverse}, the authors proposed to reverse engineer the perturbation and denoise adversarial samples with an autoencoder via class-aware alignment training.
~\cite{gandelsman2022test} leverages MAE on test-time training by reconstructing each test image at test time for corrupted samples. Inspired by this, we leverage MAE on adversarial attack detection and repair. 

\paragraph{Defense for Adversarial Attacks}
Robustness to adversarial attacks is a critical issue in the machine learning field~\cite{goodfellow2014explaining}. Large amounts of work have studied adversarial attacks in either whitebox or blackbox  (e.g., FGSM, C\&W, and ZOO attack ~\cite {goodfellow2014explaining,carlini2017towards,chen2017zoo}). To improve model robustness on these attacks, adversarial training is one of the most efficient methods~\cite{kurakin2016adversarial, madry2018towards, zhang2019theoretically, mao2019metric}. Madry et al.~\cite{madry2018towards} propose to train the model by minimizing the worst-case loss in a region around the input. TRADES~\cite{zhang2019theoretically} smoothen the decision boundary by considering the ratio between the loss of clean input and the adversarial input. Mao et al.~\cite{mao2020multitask} shows that learning multiple tasks improves adversarial robustness. However, the previously mentioned methods need retraining on the model.

Test-time defense circumvents retraining and largely reduces the computational cost. Most of the test-time adaptation adjusts the model weights with batch samples during the inference time~\cite{zhou2021domain, dou2019domain, li2018learning, zhou2020deep, wang2021tent, croce2022evaluating, gandelsman2022test }. Test-time transformation ensembling (TTE) \cite{perez2021enhancing} augments the image with a fixed set of transformations and ensemble the outputs through averaging. Test-time Training (TTT)\cite{sun2020test} trains the model with an auxiliary SSL task (e.g., rotation prediction) and adapts the model weights based on the SSL loss. MEMO\cite{memo} adapts every sample individually by minimizing the marginal cross-entropy on the transformation set. Instead of adjusting model weights, 
test-time reverse\cite{mao2021adversarial, tsai2023self} aims at adjusting input by adding some noise vectors and optimizing them with self-supervision tasks. 



\section{Method}\label{sec:method}
In this section, we define the problem and present the details of each component in the pipeline of \sys, including test-time detection and repair of attacks.


\subsection{Problem Definition}

We formalize the problem of learning to capture the natural input's manifold for detecting and repairing the malicious inputs. Assuming we have a self-supervised learning (SSL) task $\mathcal{S}$, we denote its objective function as $\mathcal{L}_s$. To capture the manifold's natural inputs for a given classifier $\mathcal{C}$, we learn the SSL task $\mathcal{S}$ using the training set of SSL task $\mathcal{S}$ doesn't involve any attacked samples.


During the inference time, given an adversarial input $u_{\rm adv}$ which is created by adding an adversarial perturbation $\delta_{\rm adv}$ to a clean sample $u\in\mathcal{Z}$, resulting in an output $\hat{y} = C(u_{\rm adv})$ that does not match the ground truth label $y$. 
To mitigate these adversarial attacks, we propose using a reverse vector $\delta_{\rm rev}$ to minimize the difference between the adapted image and the original input, effectively generating an adversarial perturbation for correction. Our goal is to find an optimal reverse vector $\delta_{\rm rev}$ with a small $\epsilon$ range that can minimize the self-supervised loss $\mathcal{L}_s$ and restore the correct prediction, such that $C(u)=y = C(u + \delta_{\rm adv}+ \delta_{\rm rev})$. 


\subsection{Learning with Masked Autoencoder (MAE)}

Self-supervised learning (SSL) helps capture the natural inputs manifold, yet standard self-supervision pretext tasks such as rotation prediction, re-colorization, and contrastive learning are not sensitive enough to adversarial attacks. The MAE is one of the strongest self-supervision learning tasks, which can repair the randomly masked-out input patches by minimizing the MAE loss of predicted patches. Motivated by this, we show how to learn the reverse vector for repairing the attacked images with MAE.

Given a masked autoencoder (MAE), we define the encoder, parameterized by $\theta$, as $f_{\theta}: \mathbb{R}^n \to \mathbb{R}^m$. Let $x \in \mathcal{X}$ be an input data in the image space, and $b \in \{0,1\}^n$ be the binary matrix that is sampled following a probability distribution. The encoder takes a masked sample $b \odot x$ as input and outputs a latent variable $z=f_\theta(b \odot x)$. We define a decoder, parameterized by $\phi$, as $g_\phi: \mathbb{R}^m \to \mathbb{R}^n$, which takes a latent variable $z$ and reconstruct the mask regions of the input image as $\hat{x}_{\rm mask} = g_\phi(z)=(g_\phi \circ f_\theta)(b \odot x)$.

The objective function of MAE is to compute the mean square error between the masked patches of input $x$ and their corresponding predicted patches, which can be defined as Eq.~\ref{eqn:mse_loss_eqn}:
\begin{align}
    \label{eqn:mse_loss_eqn}
\mathcal{L}_{\rm mse}(\cdot)=\|(1-b) \odot x - (g_{\phi} \circ f_{\theta})(b \odot x)\|_2.
\end{align}
We have discovered that the properties of the MAE can effectively capture the underlying structure of natural inputs, allowing us to achieve our objective of detecting and correcting adversarial inputs. This key insight forms an integral part of our proposed \sys method, designed to improve the robustness and reliability of machine learning systems by detecting and repairing adversarial inputs.



\subsection{Test-time Attack Detection}

After obtaining a well-trained MAE ($f_{\theta}, g_{\phi}$), the detection of anomalous adversarial attacks is straightforward by using the MAE loss $\mathcal{L}_{ mse}(\cdot)$. The rationale lies in our training for $f_{\theta}$ and $ g_{\phi}$ being on the clean sample sets $\mathcal{S}$, which can be viewed as the manifold learning on in-distribution data. Thus, the magnitude of MAE loss enlarges while encountering adversarial samples $u$ from $\mathcal{Z}$, thus triggering the detection.  

Here, we utilize MAE loss $\mathcal{L}_{ mse}$ with a statistic-based method for detection in \sys, the Kolmogorov-Smirnov test (KS test)~\cite{kstest}.
We hypothesize that loss values associated with non-attacked or clean images will be distinctly lower on average compared to those associated with attached images. Moreover, in Figure~\ref{fig:mse_loss_dist}, the sampled distributions of the loss values formed the histogram which delineates the difference in distribution between non-attacked and attacked images. Combining SSL loss with the KS test for adversarial detection is better than other detection methods that calculate the distance metric for logit outputs. In Section~\ref{subsec:detection_result}, we  illustrate the details of other baselines and demonstrate the detection results.


\subsection{Test-time Attack Repair}

The second part of the \sys method involves repairing adversarial attack inputs from malicious to benign. Unlike the existing method~\cite{gong2022reverse}, our method is fully unsupervised and can be done parallel to detection. 
To repair the potential adversarial samples $u_{adv}$, we inject a reverse vector $\delta_{rev}$ into $u_{adv}$ such that $\hat{u} = u_{adv} + \delta_{rev}$ minimizes the MAE loss $\mathcal{L}_{mse}$. We initialize the $\delta_{rev}$ under the uniform distribution following a specific $\epsilon$ range. After optimization, the adapted sample $\hat{u} = u_{adv} +\delta_{rev}$ is fed to the classifier $C$ for prediction, yielding classification scores $\hat{y} = C(\hat{u})$.
The objective function for optimizing $\delta_{rev}$ is defined as Eq.~\ref{eqn:optim}:
\begin{align}
    \label{eqn:optim}
    \delta_{rev} = \arg\min_{\|\delta_{rev}\|_{\infty} \leq \epsilon} 
    \mathcal{L}_{mse}((1-b) \odot \hat{u}, (g_{\phi} \circ f_{\theta})(b \odot \hat{u})),
    \vspace{-4mm}
\end{align} 
We aim to find an optimal vector $\delta_{rev}$ for generating a adapted sample $\hat{u} = u_{adv} + \delta_{rev}$ such that the gap between masked patches $(1-b) \odot \hat{u}$ and predicted patches $(g_{\phi} \circ f_{\theta})(b \odot \hat{u})$ of the adapted sample can be reduced. Specifically, we update $\delta_{rev}$ using projected gradient descent, subject to $\ell_{\infty}$ with a specific $\epsilon$ range. The update of $\delta_{rev}$ is described in Eq.~\ref{eqn:deltaupdate}. 
\begin{align}
    \label{eqn:deltaupdate}
    \delta^{t+1}_{rev}  = \text{clip}(\delta^{t}_{rev} + \alpha \cdot \text{sign}(\nabla_{X} \mathcal{L}_{\rm mse}), -\epsilon, \epsilon).
\end{align}
Here, $t$ is the timestamp, $\alpha$ is the step size, and $\epsilon$ is the perturbation range. The update of $\delta_{rev}$ is performed by taking the gradient of the MAE loss $\mathcal{L}_{mse}$ with respect to the input and clipping the update of $\delta_{rev}$ to the $\epsilon$ range.

\subsection{Defending against an Adaptive Attack}
\label{sec:dda_method}
We show that \sys is able to defend against an adaptive attack in which the attackers have knowledge about our defense method, including the MAE architecture, attack detection, and repair. To launch an adaptive attack that can fool both MAE and classifier $\mathcal{C}$, an attacker can generate adversarial examples by jointly optimizing the MAE loss and cross-entropy loss for the classifier. We change the optimization problem of $\delta$ to the following Eq.~\ref{eqn:optim_da} for the adaptive attack. 
\begin{multline}
    \label{eqn:optim_da}
    \text{max } 
    \mathcal{L}_{ce}(u, y), \text{s.t.} \quad 
    \mathcal{L}_{mse}((1-b) \odot u, (g_{\phi} \circ f_{\theta})(b \odot u)) \leq \epsilon
\end{multline}
        
   

The attacker should maximize the cross-entropy loss between the ground truth and output probability score of $u$ from backbone model $\mathcal{C}$, while reducing the mean-square error loss of MAE for $u$. As Eq.~\ref{eqn:constrainedoptim_da} formulates, we incorporate the Lagrange multiplier to solve the constrained optimization problem from Eq.~\ref{eqn:optim_da}.


\begin{align}
    \label{eqn:constrainedoptim_da}
    \mathcal{L}_{da}(\cdot) = 
    \mathcal{L}_{ce}(u, y)- 
    \lambda \mathcal{L}_{mse}((1-b) \odot u, (g_{\phi} \circ f_{\theta})(b \odot u)),
    \vspace{-6mm}
\end{align}
where the $\lambda$ is the parameter to control the adversarial budget. If the $\lambda$ increases, the optimization gradually shifts the budget for attacking the classification loss $\mathcal{L}_{ce}$ to the MAE loss $\mathcal{L}_{mse}$. Optimizing with $\mathcal{L}_{da}$, the optimal $\delta$ can be found to generate the defense-aware adversarial samples. We empirically found the trade-off between $\mathcal{L}_{mse}$ and $\mathcal{L}_{ce}$, where the joint optimization on  these two objective functions leads to a degradation of the attack success rate for the defense-aware adversarial samples. Our discovery is consistent with the theoretical analysis presented by Mao et al.~\cite{mao2021adversarial}. Therefore, our defense method is robust against adaptive attacks.



\section{Experiment}\label{sec:experiment}

\subsection{Experimental Settings}


\begin{table*}[t]
\centering
\small
\begin{tabular}{cccccccccc}
\hline
Detector          & FPR & FGSM-$\ell_\infty$  & FGSM-$\ell_\infty$  & PGD-$\ell_\infty$   & PGD-$\ell_\infty$   & PGD-$\ell_2$   & CW-$\ell_2$    & BIM-$\ell_\infty$   & AutoAttack \\
                  &                & 4/255 & 8/255 & 8/255 & 16/255 & 1     & 1     & 8/255 & -         \\ 
                  \hline 

FS~\cite{xu2017feature} & 0.2            &  0.001        &    0.000   &   0.0015     &  0.000     &  0.0235  & 0.022   &  0.000     &     0.000       \\
ND~\cite{hu2019new}         & 0.2            & 0.6844 & 0.6736 & 0.7024 & 0.6752  & 0.3535 & 0.6296 & 0.697     & 0.712       \\
TD~\cite{hu2019new}          & 0.2              & 0.089     & 0.067     & 0.107      & 0.125      &
0.066    & 0.051     & 0.1165     & 0.1045         \\ 
SSL-\textit{SimCLR}              & 0.2      & 0.2959      & 0.4081   & 0.6479  & 0.8163   & 0.4286   & 0.3980   & 0.6582 & 0.4948          \\
SSL-\textit{Rotation}                 & 0.2      &  0.6582     & 0.8367   & 0.9031
  &  0.9639  &  0.5510  & 0.3520  &   0.8934 & 0.7653    \\
\textbf{DRAM} (ours)           &         0.2       &   \textbf{0.8504}    &   \textbf{0.9680}    &    \
\textbf{0.9129}   &    \textbf{0.9837
}    &     \textbf{0.5900}  &   \textbf{0.5739}    &   \textbf{0.9024}    &    \textbf{0.8201}        \\  \hline
\end{tabular}
\caption{Detection results on eight kinds of adversarial attacks. We compared with five baselines, including Feature Squeezing (FS), Noise-based Detection (ND), Targeted Detection (TD), SSL detection using contrastive loss (SSL-$\textit{SimCLR}$), and loss of rotation prediction (SSL-$\textit{Rotation}$). We show the true positive rate (TPR) for every attack with the fixed False positive rate as 0.2 for all methods by varying the thresholds. The numbers in bold represent the best performance.}
\label{tab:detection_result}
\end{table*}

\newcommand{\tabincell}[2]{\begin{tabular}{@{}#1@{}}#2\end{tabular}}  

\begin{table*}[]
\small
\centering
\begin{tabular}{lccccccccc}
\hline
                           & FGSM-$\ell_\infty$ & FGSM-$\ell_\infty$ & PGD-$\ell_\infty$ & PGD-$\ell_\infty$ & PGD-$\ell_2$  & C\&W-$\ell_2$ & BIM-$\ell_\infty$  & AutoAttack \\
                         & 8/255      & 16/255     & 8/255   & 16/255  & 1  & 1   & 8/255    & -          \\ \hline
Standard   & 6.39   & 8.55   & 0.0 & 0.0 & 0.65 & 0.84 & 0.01 & 0.10          \\

+ SimCLR~\cite{chen2020simple}       & 9.46      & 8.82     & 1.89     & 0.12   & 15.59  & 17.62       & 1.74    & 3.31          \\
+ Rotation~\cite{gidaris2018unsupervised}      & 11.40      & 9.85     & 1.47     & 0.98  &  0.64   & 1.02    & 3.89    & 1.49              \\

\textbf{+ \sys (ours)}               & \textbf{20.54}      & \textbf{14.14}      & \textbf{21.34}   & 
\textbf{13.12}   & \textbf{39.57}    & \textbf{41.79}    & \textbf{20.51}    & \textbf{22.4}          \\

\hline

MADRY~\cite{madry2017towards}     & 24.95      & 13.9      & 20.38   & 8.83   & 35.11    & 40.82    & 20.26    & 16.08          \\

+ SimCLR~\cite{chen2020simple}              & 24.19     & 13.31     & 20.33   & 8.75  & 35.80    & 43.37    & 20.31    & 18.84          \\
+ Rotation~\cite{gidaris2018unsupervised}                 & 22.57     & 12.35     & 19.94   & 8.94  & 34.38    & 41.00    & 20.04    & 22.69          \\
\textbf{+ \sys (ours)}                & \textbf{25.43}      & \textbf{14.05}      & \textbf{23.21}   & \textbf{11.01}   & \textbf{38.71}   & \textbf{43.75}    & \textbf{23.14}    & \textbf{24.84}          \\

\hline

\end{tabular}
\caption{The repair results on eight kinds of attack. We compare \sys with the other two SSL baselines, including contrastive learning task (SimCLR~\cite{simclr}) and rotation prediction~\cite{gidaris2018unsupervised}. We show the robust accuracy (\%) on two Resnet50~\cite{he2016deep} models pre-trained with Standard training and $\ell_\infty$ robust training (MADRY~\cite{madry2017towards}). The numbers in bold represent the best performance.}
\label{tab:reconstruction}
\end{table*}

\paragraph{Dataset Detail}
We train and test \sys using the ImageNet dataset~\cite{deng2009imagenet}, a 1K-class large-scale vision dataset. Notice that the training process of \sys uses only the original ImageNet train set with $\sim$1M images.
To evaluate \sys, we generate the adversarial samples on the ImageNet test set~\cite{deng2009imagenet} with 50K images. Here, we consider eight attacks and generate them from two kinds of models, the  standard ~\cite{he2016deep} and robust~\cite{madry2017towards} ResNet50. 

\paragraph{Model Detail}
The MAE model is based on the visual transformer (ViT) arcitecture~\cite{Visualtransformer}, which consists of the same encoder module as ViT that splits images into patch size 14$\times$14 and encodes their corresponding projected sequences. The encoder module of MAE consists of alternating layers of multiheaded self-attention (MSA) and MLP blocks. The batch normalization layer (BN) is applied before every block, and residual connections after every block. The encoded features are further processed by the decoder module for reconstruction.

\subsection{Test-time Attack Detection}
\label{subsec:detection_result}

We compare the detection part of \sys with the following four kinds of baselines:

$\bullet$ \textbf{Feature Squeezing (FS)}: The baseline~\cite{xu2017feature} utilizes multiple data transformations such as median smoothing, bit quantization, or non-local mean to extract the squeezing features of input samples. The $\ell_1$ distance is then computed between the transformed version and the original version of inputs for detection. If the $\ell_1$ distance is larger than the threshold, the samples are rejected as adversarial.

$\bullet$ \textbf{Noise-based Detection (ND)}: This method~\cite{hu2019new} detects whether an input is adversarial or not via perturbing the input with random noise. It samples the $\epsilon$ in a normal distribution with specific radius $r$ and computes the $\ell_1$ distance as $\parallel h(x) - h(x+ \epsilon) \parallel_1$ for given input $x$. The input $x$ is detected as adversarial if the $\ell_1$ distance is large.

$\bullet$ \textbf{Targeted Detection (TD)}: The detection method~\cite{hu2019new} aims to evaluate the PGD attack algorithm on every input $x$ and record the number of steps $K$ to fool the sample $x$ from the current predicted class to targeted class. If the number of steps $K$ is larger than the threshold, we detect the sample $x$ as adversarial.

$\bullet$ \textbf{Detection with Self-Supervised Loss (SSL)}: The detection method combines SSL loss with the 2-sample KS-test. It measures the distributions of the observations from the clean and attacked samples. The samples are rejected as adversarial if the p-value exceeds a threshold. We compare \sys with two SSL tasks, including contrastive learning (SimCLR~\cite{simclr}) and rotation prediction~\cite{gidaris2018unsupervised}.

\begin{figure*}[t]
\centering
\includegraphics[width=0.95\linewidth]{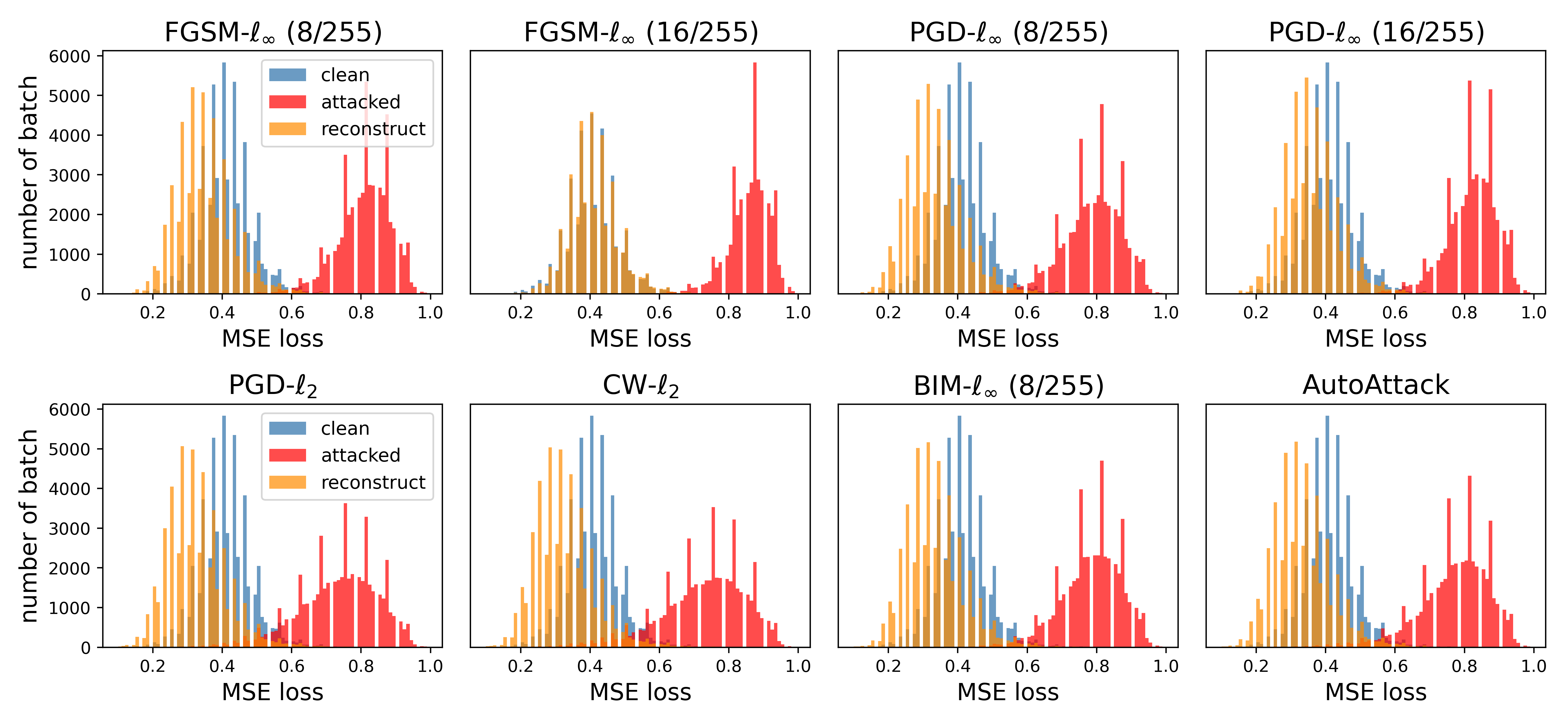}
\vspace{-4mm}
         \caption{Histogram plot of self-supervised losses $\mathcal{L}_{mse}$ for eight adversarial attacks. We show the loss distribution of original images with the color blue, adversarial images as red, and adapted images as green. The figure shows that adversarial images cause a large shift in MAE loss (from blue to red). After using \sys to repair, the loss distribution of adapted images shifts back and gets closer to the original images (red to yellow).}
             \label{fig:mse_loss_dist}
    \vspace{-2mm}
\end{figure*}



Table~\ref{tab:detection_result} shows the true positive rate 
(TPR) of our method against eight kinds of adversarial attacks on ImageNet data as the detection rate. Our method combines the $\mathcal{L}_{mse}$ loss with a two-sample KS test and set the p-value threshold based on a fixed false positive rate (FPR) of 0.2. Here, we evaluate the detectors for every baseline using the same FPR.
Compared with feature squeezing~\cite{xu2017feature}, noise-based detection~\cite{hu2019new}, targeted detection~\cite{hu2019new}, and detection with the two SSL tasks, our detection rate hovers around 82\%, which outperforming all other baselines. The superior detection results of \sys suggest that the MAE loss is more sensitive to variance in distributions. Compared to the other two self-supervision tasks, \sys outperforms rotation prediction by up to 20\% and contrastive learning by up to 60\%. Moreover, the two-sample KS-test has the additional benefit of being able to better control the TPR-FPR tradeoff compared to all other baselines by varying the p-value threshold for class prediction. 

\begin{figure}[t]
\centering
\includegraphics[scale=0.51]{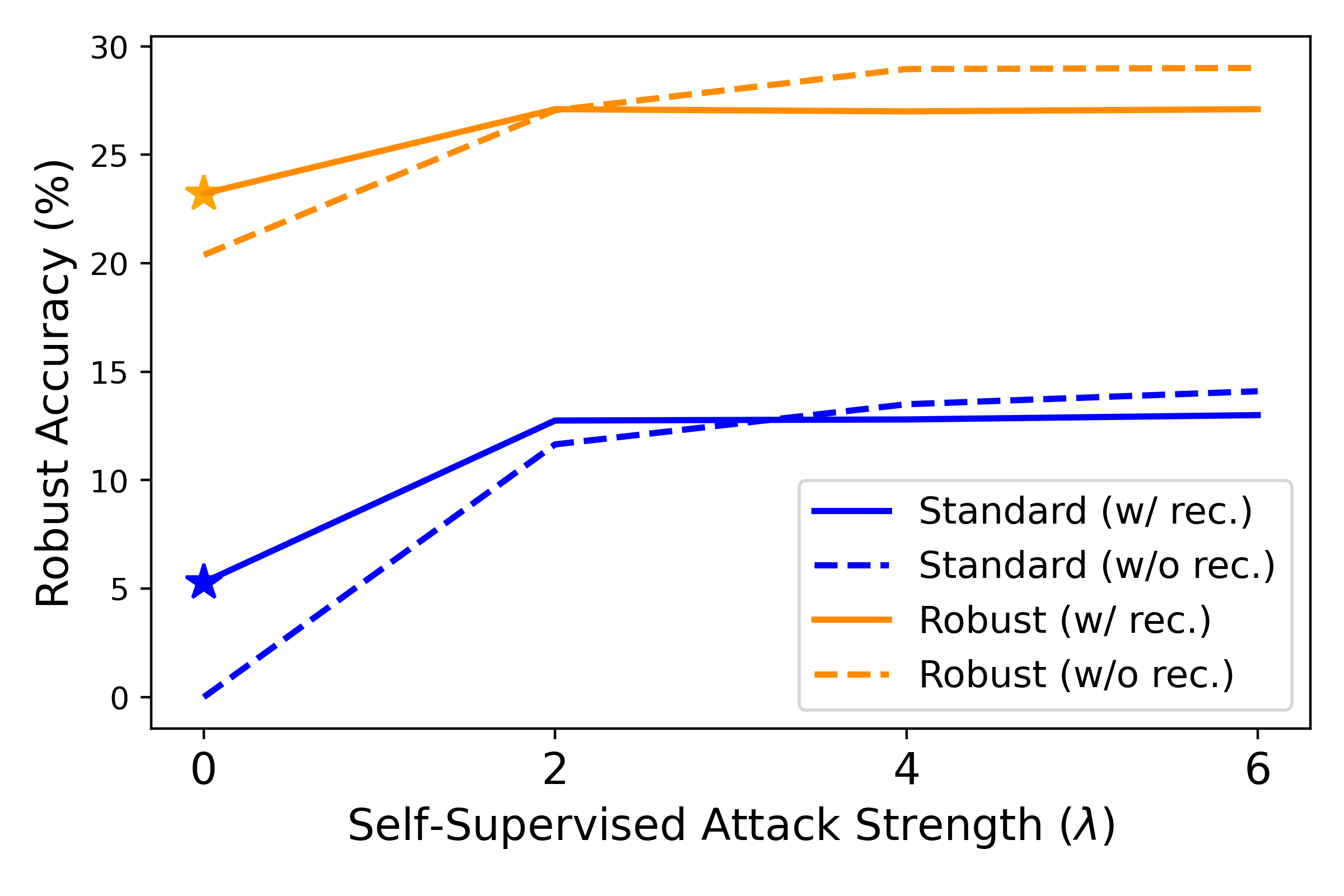}
\vspace{-3mm}
         \caption{Defense-aware Attack: We show the trade-Off on the robust accuracy and SSL attack strength. When increasing the SSL attack strength $\lambda$, the robust accuracy increases, which means the attack is less effective. When the $\lambda$ is set as zero, \sys has the best performance on robust accuracy. }
\label{fig:daa_robust_acc}

\end{figure}



\subsection{Attack Repair Results}
We compare our repair method with two kinds of SSL baselines (1) Contrastive learning learns the representations that map the feature of the transformations of the same image into a nearby place. We follow the SimCLR~\cite{simclr} to implement the contrastive learning task. We create positive/negative pairs for images by applying some transformations such as random cropping/flipping and random rotation. (2) The rotation prediction~\cite{gidaris2018unsupervised} learns to predict the rotation degree of upcoming input samples. Here, we train the rotation prediction task as one of our SSL baselines with 4 classes ($0^{\circ}$, $90^{\circ}$,$180^{\circ}$, $270^{\circ}$).

As Section~\ref{sec:method} mentioned, we optimize the perturbations for a given input by using the PGD algorithm based on the MAE loss $\mathcal{L}_{\rm mse}$ extracted from MAE. Table~\ref{tab:reconstruction} shows we evaluate the robust accuracy on the target classifiers $\mathcal{C}$ from which our adversarial samples were generated. \sys successfully repairs the adversarial images and improves the robust accuracy without requiring the ground truth labels at the inference time. The target classifiers are pre-trained with two different methods, including standard training and robust training.
As Table~\ref{tab:reconstruction} shows, 
compared with the two baselines using contrastive and rotation losses, \sys improves the robust accuracy after repair by $6\%\sim40\%$ for the attack samples generated from the standard ResNet50 model, whereas contrastive learning (SimCLR) and rotation prediction improves only by 1\%$\sim$7\%. For the Robust ResNet50 model, \sys improves the robust accuracy by 1\%$\sim$8\% for every attack, outperforming the two baselines.



\subsection{Defense-aware Attack (DAA) Results}
As Section~\ref{sec:dda_method} mentioned, we launch the adaptive attack on models and demonstrate the effectiveness of \sys on this stronger attack which assumes the attacker knows our defense method and the MAE model. Figure~\ref{fig:daa_robust_acc} shows the result of the defense-aware attack on two classifiers, including standard ResNet50 and robust ResNet50~\cite{madry2017towards}. While increasing the self-supervised attack strength $\lambda$ for DAA from 0 to 6, the attacks are gradually weakened, and the robust accuracy increases. If the adversary attempts to attack the MAE, the attack will be less effective on the backbone classifiers. On the other hand, if the adversary attempts to attack the classifier by generating samples with high attack success rates, the MAE loss can't be minimized together. While the dotted line shown in Figure~\ref{fig:daa_robust_acc} indicates the result of DAA, the solid line shows the repair result of \sys on top of DAA, which indicates that our method has optimal results when $\lambda$ is set as 0 for both the standard and robust model. Therefore, it is difficult for adversaries to attack both MAE and classifiers at the same time. 

\vspace{-2mm}

\subsection{Loss Analysis on Attack Detection}
\label{ablation:loss}
To better understand the ability of repair for MAE, we visualize the histogram on the self-supervised loss distribution for adversarial examples.  We compare the loss variance between clean samples and different kinds of attack samples by collecting the loss on every batch during the testing-time inference. As Figure~\ref{fig:mse_loss_dist} shows, the loss variance on clean samples (colored with blue region) and the other eight attack types' samples (red curve) are large before repair. The orange curve shows the loss distribution of samples after repair. Here, \sys uses MAE loss to repair the samples and demonstrates a huge loss shifting after repair (colored with yellow region), which is closer to the loss distribution of clean samples. This result inspires us to leverage the MAE loss to detect the adversarial sample with KS-test.

\section{Conclusion}\label{sec:conclusion}
We have presented a novel test-time defense called DRAM against (previously unknown) adversarial attacks. By using the self-supervision task MAE, \sys effectively detects and repairs the adversarial samples during the inference time. Results show that \sys achieves an average 82\% detection score on eight previously unseen attacks. Our method for attack repair improved the robust accuracy of ResNet50 by up to 40\% and 8\% under standard and robust training after repair. In addition, we show \sys is possible to defend against a defense-aware attack and delineate the trade-off between the attack budgets of the self-supervision task and the classification task. Overall, \sys improves the robust accuracy across multi-type of adversarial attacks at the inference time, which provides new insight into the frontier of test-time defense for unseen attacks.

\section{Acknowledgement}\label{sec:acknowledge}


This work is partially supported by a GE / DARPA grant (Agreement \# HR00112090133), a CAIT grant, and gifts from JP Morgan, DiDi, and Accenture.

{\small
\bibliographystyle{ieee_fullname}
\bibliography{PaperForReview}
}

\end{document}